\documentclass[accepted]{uai2023} 

\usepackage[american]{babel}

\usepackage{natbib} 
\usepackage{mathtools} 
\usepackage{booktabs} 
\usepackage{tikz} 
\usepackage{amsmath}
\usepackage{graphicx}
\usepackage{subcaption}

\title{Heteroskedastic Geospatial Tracking with Distributed Camera Networks}

\author[1]{\href{mailto:<csamplawski@cs.umass.edu>?Subject=Your UAI 2023 paper}{Colin~Samplawski}{}}
\author[1]{Shiwei Fang}
\author[2]{Ziqi Wang}
\author[1]{Deepak Ganesan}
\author[2]{Mani Srivastava}
\author[1]{Benjamin~M.~Marlin}
\affil[1]{%
    Manning College of Information \& Computer Sciences\\
    University of Massachusetts Amherst
}
\affil[2]{%
   University of California, Los Angeles
}
  
\begin{document}
\maketitle

\begin{abstract}

Visual object tracking has seen significant progress in recent years. However, the vast majority of this work focuses on tracking objects within the image plane of a single camera and ignores the uncertainty associated with predicted object locations. In this work, we focus on the geospatial object tracking problem using data from a distributed camera network. The goal is to predict an object's track in geospatial coordinates along with uncertainty over the object's location while respecting communication constraints that prohibit centralizing raw image data. We present a novel single-object geospatial tracking data set that includes high-accuracy ground truth object locations and video data from a network of four cameras. We present a modeling framework for addressing this task including a novel backbone model and explore how uncertainty calibration and fine-tuning through a differentiable tracker affect performance. 
\end{abstract} 

\section{Introduction}\label{sec:intro}
Deep neural network models have enabled remarkable advances in visual object tracking performance across a wide range of scenarios over the last decade \citep{survey}. The task of visual object tracking involves detecting and localizing objects as they move through a scene. The vast majority of prior work on visual object tracking considers video data from a single camera only and poses the problem as tracking objects within the image plane of the camera. Further, most approaches to tracking provide outputs as a sequence of bounding boxes that lack a representation of uncertainty over the locations of object.

In this work, we focus on the geospatial object tracking problem using a distributed camera network. The goal is to predict an object's track in geospatial coordinates along with uncertainty over the location while accounting for communication constraints that prohibit centralizing raw image data. This problem is motivated by the proliferation of Internet of Things (IoT) devices with local sensing, compute and wireless communication capabilities. The geospatial object tracking problem has important applications in smart cities such as traffic monitoring and pedestrian safety that require knowledge of where objects are in real-world map coordinates \citep{datta2017survey}.

Importantly, the continuous deployment of camera and other sensor networks in real-world environments introduces multiple challenges due to resource constraints and changing environmental conditions \citep{pereira2020challenges}. Due to resource constraints the set of available cameras may not cover the entire environment of interest resulting in observability gaps. Objects may occlude each other from some vantage points but not others as they move through the environment. Lighting and weather changes will also effect some or all cameras at different times. As a result, it is essential to develop models and systems that can refelect the uncertainty in an object's location in a meaningful way taking into account local observability, occlusions, and environmental effects. These aspects are often overlooked in traditional tracking benchmarks that assume access to high-quality video data \citep{MOTChallenge20}.



To facilitate research on this problem, we collected a novel single-object tracking data set that includes high-quality ground truth geospatial object locations collected at 100 samples per second using a motion capture system. These data are combined with video from four camera nodes with different partially overlapping views of the environment. The object tracked is a remote controlled vehicle. The dataset spans multiple scenarios including an open environment and an environment with occluding structures, as well as a normal and low lighting scenarios. This data set and associated processing algorithms  will be released along with the paper. 


We present a modeling framework for addressing this task that augments neural network object detection backbones with adapters that translate within image-plane features into geospatial coordinates with associated uncertainty. We refer to such as model as a heteroskedastic geospatial detector (HGD). To address communication constraints, each HGD is restricted to operate over data available locally at a singe camera node. To solve the tracking problem, the low-dimensional probabilistic outputs of a set of independent HGDs are centralized and fused together using a multi-observation Kalman filter model \citep{kalman}. 

We experiment with both existing and custom backbones within the HGDs. Further, we explore the effects of geospatial detection calibration to improve the quality of the distributions provided by individual HGDs to the Kalman filter. Finally, we experiment with backpropagating loss through the Kalman filter and in to individual HGD adapters as a form of fine tuning and assess its impact on tracking performance.

Our results show that re-calibrating the output of the HGD models using an affine transformation of their raw covariance outputs can significantly improve log likelihood of true object locations. We show that fine tuning by backpropagating through the Kalman filter further improves performance for most HGD models. 

Finally, we show that our custom HGD backbone can provide performance on par with ResNet50 and DETR backbones under the normal lighting condition while simultaneously providing significant improvements in run time latency. Interestingly, we show that in the low-light condition, the DETR-based model has remarkably more robust performance than the other models considered. 



The rest of the paper is organized as follows:
Section \ref{sec:related_work} discusses prior work on tracking. Section \ref{sec:dataset} presents the data set including the data collection process and subsequent data post-processing. Section \ref{sec:model} outlines our modeling framework. Section \ref{sec:experiments} presents our experimental protocol and evaluation metrics. Section \ref{sec:results} presents results. We conclude with a discussion in Section \ref{sec:conclusions}.

\section{Related Work}\label{sec:related_work}
There is considerable prior work applying deep learning to visual object detection and tracking problems. \cite{survey} present a survey paper that discusses many of these approaches in depth. In this section we highlight a handful of papers that are of particular relevance to our work, noting that we focus on the multi-view geospatial tracking problem and not the traditional single-view, within-image-plane tracking problem.

\textbf{Model Architectures:} Convolutional neural networks \cite{krizhevsky2017imagenet} provide the backbone in most deep learning object detection and tracking models. In this work, we experiment with a ResNet50 backbone as an example of this class of approach \citep{resnet50}. More recently, vision transformer-based models have shown promising performance on image classification \citep{swin}, object detection \citep{detr}, and within-image-plane object tracking \citep{trackformer,motr,yan2021learning}. The attention mechanism used in the transformer architecture of \cite{vaswani2017attention} has proven to be a powerful tool for reasoning about track identity through time. We consider backbone models based on two transformer architectures. 

\textbf{Multi-view Computer Vision:} The task of fusing data from multiple cameras to solve computer vision problems has a long history. One popular application is the use of multiple views to enable 3D object detection in autonomous driving systems, as seen in recent works such as DETR3D \citep{detr3d} and \cite{chen2017multi}. While some approaches predict object location in an overhead "bird's-eye-view" space, such as \cite{Can_2021_ICCV}, these approaches still rely on predicting bounding boxes and do not account for object location uncertainty. More recently, the work of \citep{bevformer} introduced a novel transformer architecture to fuse multiple camera views to make "bird's-eye-view" predictions. However, this method assumes that camera data can be centralized at no cost and thus relies on early fusion while we focus on the communication constrained distributed camera network setting.

\textbf{Probabilistic Tracking:} Many early approaches to tracking leverage probabilistic models \citep{perez2002color}. The Kalman filter is a commonly used approach due to its simplicity \citep{kalman}. For example, the Simple Online and Realtime Tracking (SORT) approach of \cite{SORT} converts bounding boxes predicted from an object detector into a track using a Kalman Filter. However, this approach does not make probabilistic predictions for tracked objects and operates within the image plane of a single camera. \cite{Danelljan_2020_CVPR} pose the tracking problem as a probabilistic regression that minimizes KL divergence. However, unlike our approach they consider ground-truth locations that are represented as bounding boxes where as we represent distributions over the centroid of a tracked object.
\begin{figure*}[htbp]
    \centering
    \includegraphics[width=\textwidth]{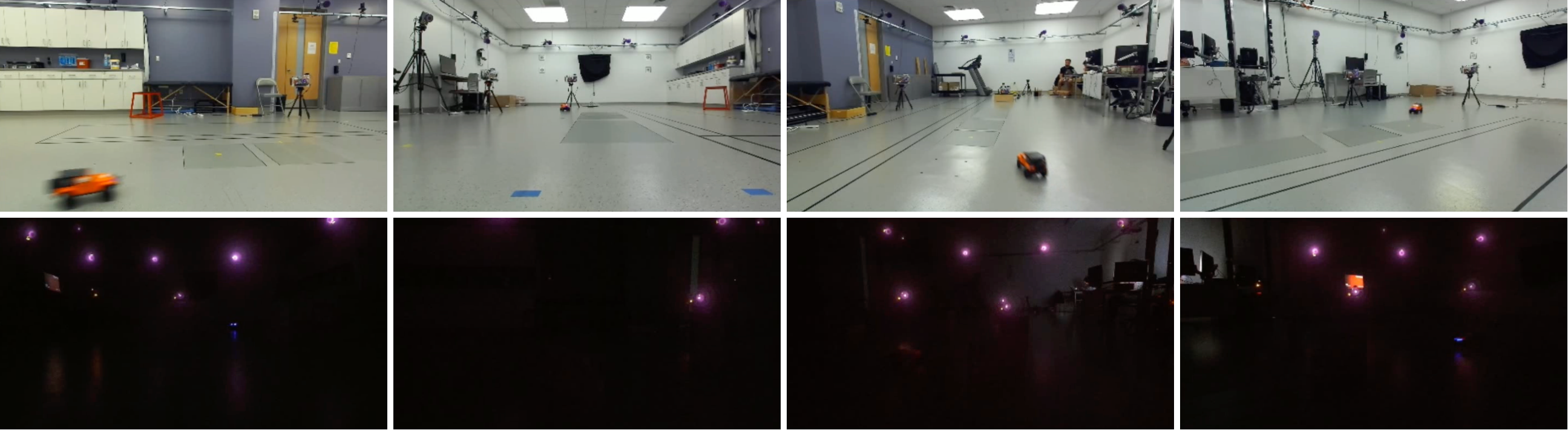}
    \caption{Example data from each of the four cameras in the normal lighting (top) and low lighting scenarios (bottom).}
    \label{fig:data_example}
\end{figure*}

\vspace{-1em}
\section{Geospatial Tracking Dataset}\label{sec:dataset}
In this section we present the data collection process and data pre-processing methods used to create our data set.

\textbf{Data Collection Infrastructure:} The data set was collected using RGB video data obtained from a network of four camera nodes. Each camera node included a $1080 \times 1920$ resolution ZED 2i camera and a Jetson Xavier NX providing local video compression and storage. The data collection experiments were performed in an indoor motion capture environment measuring $5\times 7$ meters in size. The motion capture system provides  high-quality ground truth location and orientation data for tracked objects at a sampling rate of 100Hz. We used a single remote control vehicle (an orange truck) as the tracked object in all data collection experiments. This object is $15 \times 30$ cm in size. 

The four camera nodes were located on the four sides of the rectangular motion capture environment looking inwards. The ZED camera has a 2.1mm lens, producing a 120-degree field of view. The nodes were positioned such that their fields of view overlapped while no single node captured the whole environment. This allows the tracked object to be out of the field of view of different camera nodes at different times. The locations and orientations of the camera nodes remained constant throughout all data collection experiments.



\textbf{Tracking Scenarios:} Data were collected under two different scenarios. In the first scenario the environment is open (no occluding objects are present) and fully lit. In the second data collection scenario, the environment is minimally lit and occluding objects are also present. While the tracked object has onboard illumination (headlights) that make tracking it plausible under low lighting, the second scenario is expected to be significantly more difficult than the first scenario. 
Examples from both scenarios are displayed in Figure \ref{fig:data_example}. Each data scenario has a total length of 5 minutes. We split each scenario such that 2.5 minutes are used for training, 30 seconds are used for validation, and 2 minutes are used for testing. Video data are recorded at 15 FPS.

\textbf{Data Pre-processing:} We construct a multi-view tracking data set as a sequence of instances where each instance contains four images (one from each of the four camera nodes) and the corresponding ground truth object location. Given that the cameras are operating independently on the four nodes, we deal with video stream synchronization disparities by initializing a buffer for each video stream as well as for the the  ground truth object location stream output by the motion capture system. We update these buffers as the data arrives from each source. We create individual instances in the tracking sequence by taking a snapshot of all buffers at a rate of 20 frames per second. Before learning models, we downsample the image data by dividing each side length by a factor of 4. This results in input images that have size $270 \times 480$. No further data augmentation is applied. 

\begin{figure*}[t]
     \centering
     \begin{subfigure}[b]{0.48\textwidth}
            \centering
            \includegraphics[width=\columnwidth]{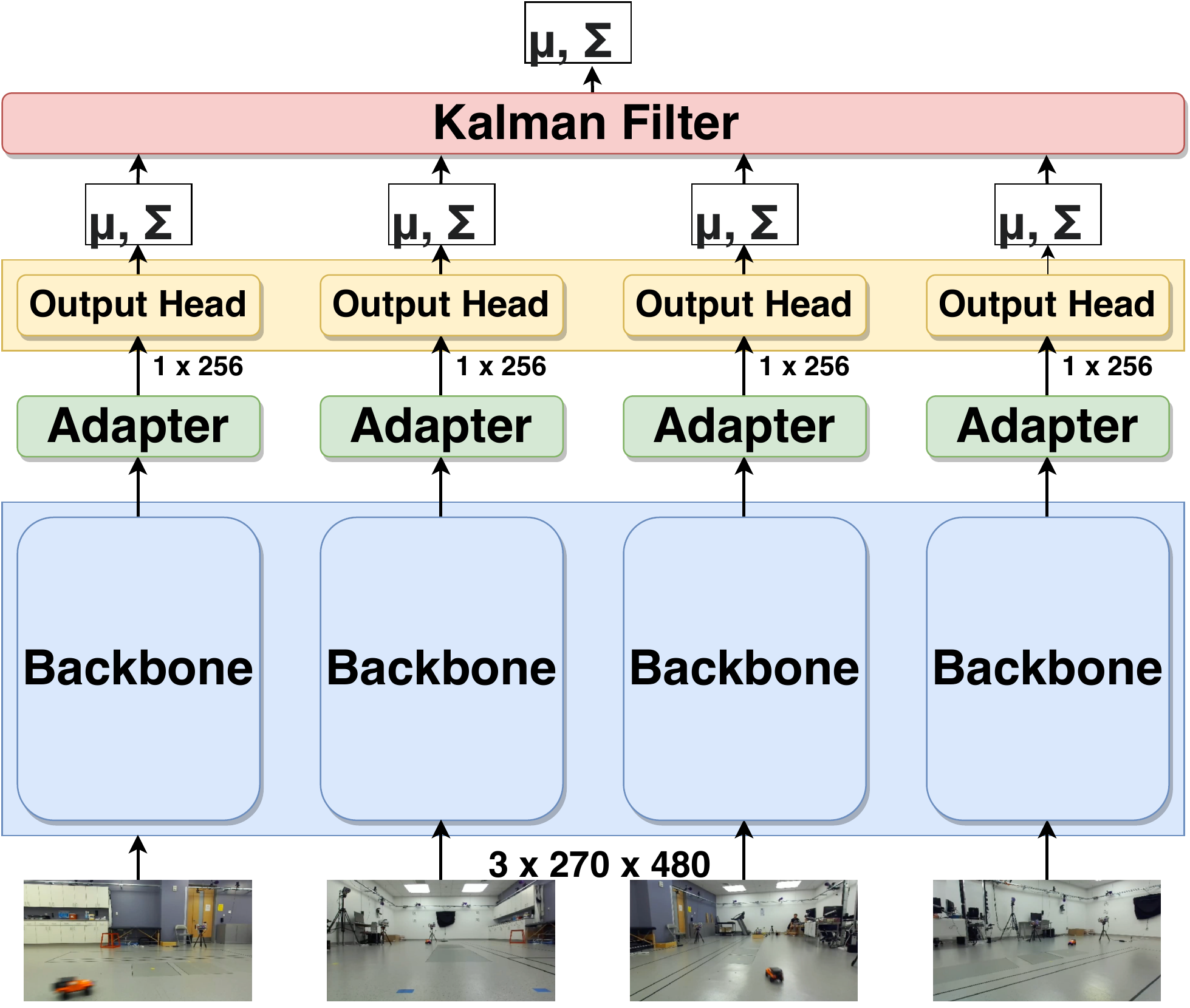}
            \caption{}
            \label{fig:general_model}
     \end{subfigure}
     \hfill
     \begin{subfigure}[b]{0.48\textwidth}
            \centering
            \includegraphics[width=\columnwidth]{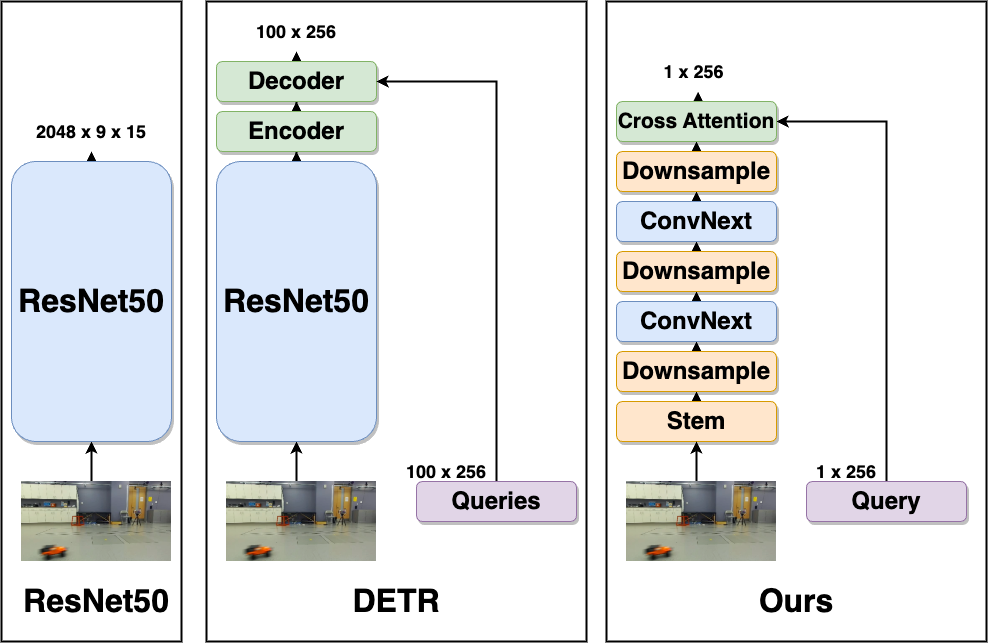}
            \caption{}
            \label{fig:backbones}
     \end{subfigure}
        \caption{(a) General model architecture. Layers with rectangular background denote model components which are shared across views. 
        (b) Backbone model block diagrams. }
\end{figure*}


\section{Models for Heteroskedastic Geospatial Tracking}\label{sec:model}

In this section we present our proposed heteroskedastic geospatial tracking model framework. The framework consists of two primary components: a set of heteroskedastic geospatial detection (HGD) models and a geospatial tracker (GST) based on a multi-observation Kalman filter.

\textbf{Overview:} An HGD model takes an image from a single view as input and predicts a full covariance normal distribution over the location of the object in geospatial coordinates. Specifically, an HGD model outputs a 2-dimensional mean object location $\mu$ and a $2 \times 2$ object location covariance matrix $\Sigma$. There is one HGD model for each view and all HGD models operate independently. The GST component then filters a set of HGD model outputs in order to generate smoother trajectories that fuse the predictions from each view taking into account their associated object location uncertainties. A block diagram illustrating this framework is presented in Figure \ref{fig:general_model}.

\textbf{HGD Models:} An HGD model is further divided into three sub-components: a backbone, an adapter, and an output head. The backbone processes the input images in order to generate general higher-level features. The adapter is a small feed-forward network which converts the features into a single $256$-dimensional vector which we treat as a latent encoding of the object's position in geospatial coordinates. Finally, the output head converts this encoding into the final low-dimensional geospatial object location mean and covariance representation using an additional feed-forward network. We share the parameters of the backbone and output head across all views. We instantiate a separate adapter for each view. This allows the general backbone features to be translated into view-specific features. 

\textbf{HGD Backbones:} We provide results using two baseline backbone models: ResNet50 \citep{resnet50} and DETR \citep{detr} as well as a custom backbone. A visual description of the three backbone models is shown in Figure \ref{fig:backbones}. 

ResNet50 is a popular CNN model that was originally designed for image classification \citep{resnet50}. We tap a pretrained model at the last feature layer. This model generates a feature map with height and width downscaled by a factor of 32 relative to the input. The feature map has 2048 channels. The HGD adapter converts the feature map to $256$ channels via a feed forward network and then flattens out the spatial dimensions using a linear projection. 

DETR is a recent object detector which performs detection end-to-end using transformers \citep{detr}. It consists of a ResNet50 model followed by a transformer encoder-decoder pair. Using self-attention and cross-attention the model learns a set of 100 query embeddings which act as latent encodings of the predictions. We use the output of the last decoder layer which has shape $100 \times 256$. The adapter applies a feed forward network and then applies a linear layer to project the 100 query embeddings to a single prediction (i.e. $1 \times 256$). We note that DETR was originally trained for within-image-plane bounding box object detection. Therefore, the adapter has the additional duty of translating this information into the geospatial coordinate space. 

Lastly, we introduce an additional custom backbone model for this task. The general design is inspired by DETR, while being much more lightweight. We start with a custom CNN feature extractor. First we copy the stem of a ResNet50 model, which consists of the first convolution layer of the model and a max pooling layer. This stem was trained on ImageNet and is left frozen throughout training and evaluation. We then interleave ConvNext blocks  and downsampling operations. The ConvNext block \citep{convnext} is a simple residual CNN block consisting of a depthwise and pointwise convolutional layer. 
Downsampling is done via convolutional layers with stride 2. In total we apply 3 downsampling layers and 2 ConvNext layers. Along with the stem, this results in just 8 convolutional layers. We maintain a feature dimension of 256 throughout the CNN and output a feature map with height and width downscaled by a factor of 32 relative to the input image.

Associated with this backbone is a single query embedding with 256 dimensions. It is treated as a vector of free parameters. We take the output of the CNN stage and flatten the spatial dimensions into a sequence of pixel values. We then apply cross attention between the query embedding and this pixel sequence. This has the effect of compressing the visual information from the CNN input into a single 256 dimensional vector. This is then fed into an adapter layer.

We present a runtime analysis of these three models in Table \ref{runtime}. We report only the latency of the backbone models without taking into account the latency of the Kalman filter, which is constant for each model and trivial compared to the latency of the deep models. The latencies reported are the time taken to generate a predicted distribution for every view. We see that our custom model is 6 times faster than DETR and 2.5 faster than ResNet50. Run time latencies were measured using an NVIDIA 3080TI GPU.

\begin{table}[h]
\centering
\caption{Runtime Results}
\label{runtime}
\begin{tabular}{lrr}
\toprule
   Model &  Latency (ms) &    FPS \\
\midrule
ResNet50 &         21.23 &  47.12 \\
    DETR &         51.02 &  19.60 \\
    Ours &          8.05 & 124.25 \\
\bottomrule
\end{tabular}
\end{table}

\textbf{HGD Output Head:} The HGD output head is responsible for predicting the mean $\mu$ and covariance matrix $\Sigma$ of the Gaussian distribution that represents the location of the object. Given a 256-dimensional encoding, it outputs five values. The first two values determine the mean location $\mu$. In our experiments, we apply a sigmoid activation to these values and then scale them by 500 and 700 so that they correspond to a location in centimeters within the tracking environment's  5$\times$ 7 meter area. 

The remaining three output values are used to predict the covariance matrix $\Sigma$. We apply a softplus function to two of the values which are taken to be the diagonal elements of $\Sigma$. The remaining value is then taken to be the off-diagonal element. Finally, we multiply this intermediate matrix with its transpose to ensure that $\Sigma$ is positive definite.  For numerical stability reasons, we add the identity matrix $I$ to the output. This prevents $\Sigma$ from collapsing to $0$ during learning.

\textbf{GST Model:} For the GST model, we use a constant-velocity multi-observation Kalman filter-based tracker \cite{kalman,baisa2020derivation}. A Kalman filter-based tracker is a dynamic Bayesian network model with Gaussian distributed latent state. In the constant velocity Kalman tracker, these latent variables represent the location and velocity of the object being tracked. A detection is modeled as a noisy observation given the latent variables in the tracker. Posterior inference is used to update the tracker's latent state given a detection. 

\begin{figure*}[t]
\centering
\begin{subfigure}{.45\textwidth}
  \centering
  \includegraphics[width=\linewidth]{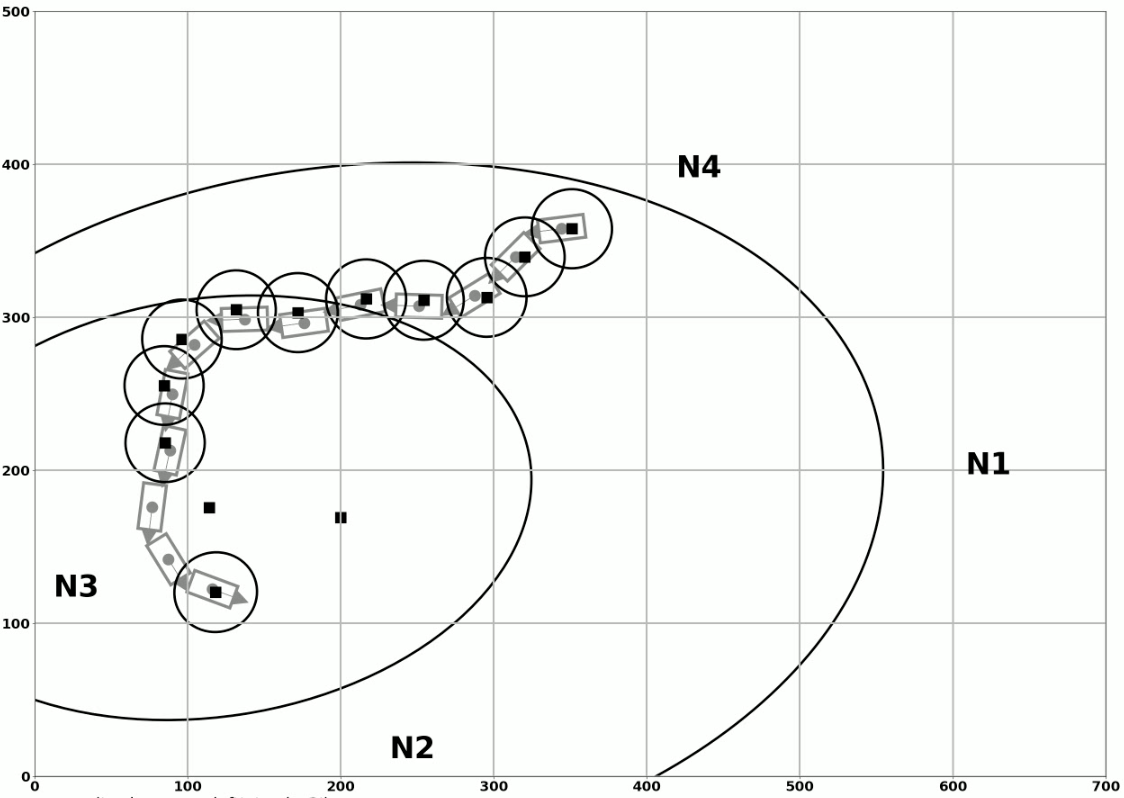}
  \caption{Example trajectory from single view detector.}
\end{subfigure}\hfill %
\begin{subfigure}{.45\textwidth}
  \centering
  \includegraphics[width=\linewidth]{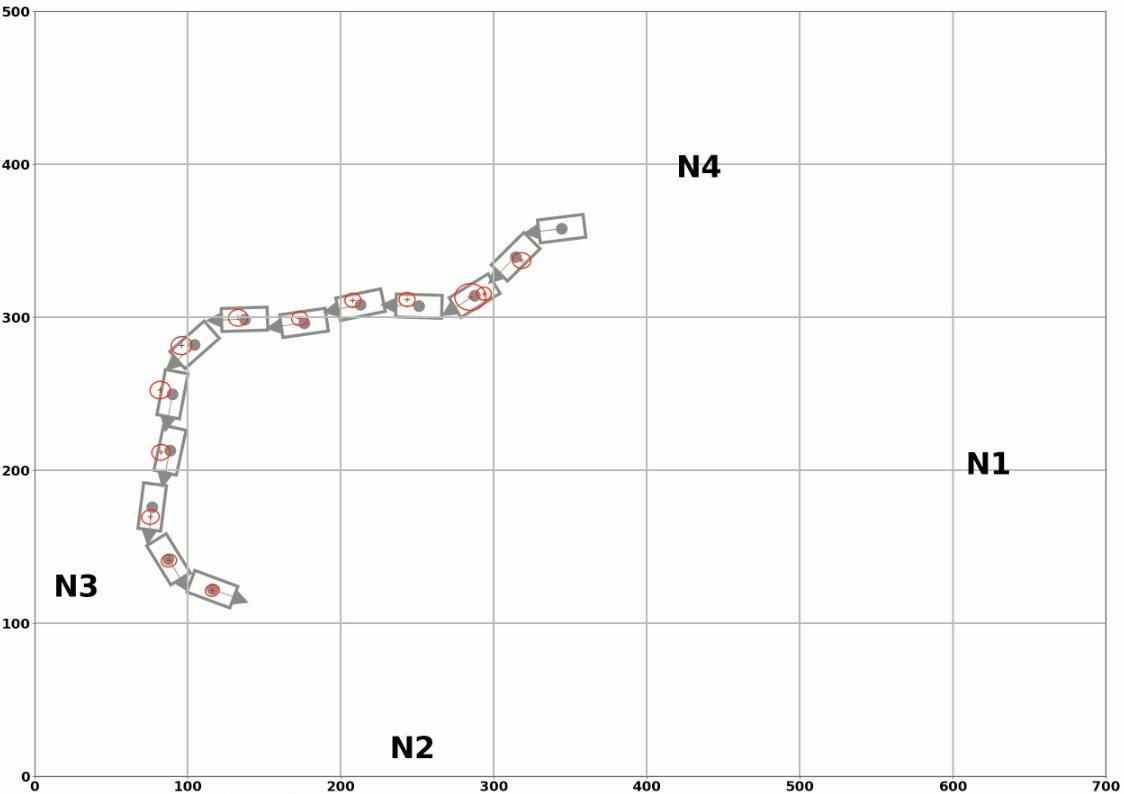}
    \caption{Example trajectory from multi-view tracker.}
\end{subfigure}
\caption{Example trajectories. The gray rectangles denote the location of the object being tracked with an arrow indicating the object's current heading. The location of the camera nodes are labeled as N1 to N4. The left-hand plot shows detections from the custom backbone using data from node N2. The black squares show the predicted mean location with an associated 95\% confidence ellipse. The right-hand plot shows the same track with multi-observation Kalman Filter output shown in red. We see that for two time points the single-view detector predicts a mean that is off track with high uncertainty. This occurs when the object is not viewable by node N2. However,  the tracker remains on tracker as it fuses data from the other nodes that can see the object and have much higher confidence.}
\label{fig:example_trajectories}
\end{figure*}

In a multi-observation Kalman filter, any number of simultaneous detections can be used to update the latent state of the tracker under the assumption that they are all mutually conditionally independent given the latent state of the tracker. The posterior inference process fuses multiple simultaneous observations while automatically putting more weight on the detections that have lower associated uncertainty. An example of this effect can be seen in Figure \ref{fig:example_trajectories}.



\textbf{HGD and GST Training:} We train the HGD models to minimize of the negative log likelihood of the ground-truth object positions under the distribution output by the model. All views are trained simultaneously with a separate loss contribution for each view. All models are trained for 50 epochs. We use the AdamW optimizer \citep{adamw} with an initial learning rate of $10^{-4}$ and weight decay of $10^{-4}$. At the 40th epoch, the learning rate is divided by 10. We train on 8 GPUs each with 4 samples per batch for an effective batch size of 32. Batches are chosen randomly across the training sequence as temporal information is not used during training of the HGD models. The gradient is clipped such that it's L2 norm is less than 0.1. We start from a ResNet 50 model pretrained on ImageNet \citep{deng2009imagenet} and a DETR model pretrained on the COCO detection dataset \citep{lin2014microsoft}. Parameters of the baseline models are fine-tuned during training of the detector. 

In the experiments where we backprop through the Kalman Filter, a batch size of 1 is used on each GPU for an effective batch size of 8. We use time aligned sequences of length 100 for this portion of training. When training with the Kalman Filter we use detectors that are already trained on the task. The backbones are frozen during the Kalman Filter training while the paramaeters of the Kalman Filter, output head and adapters are fine-tuned. All learning hyperparameters are the same for GST and HGD training except for the training schedule, which is decreased in length by a factor of 10 (we train for 5 epochs and reduce learning rate at 4th epoch).

\section{Evaluation Metrics}\label{sec:experiments}

In this section we present metrics for evaluating the performance of models that solve the heteroskedastic geospatial object tracking problem.

\begin{table*}[t]
\centering
\caption{Baseline tracking results}
\label{lightsT_baseline}
\begin{tabular}{l|rrrr|rrrr}
\toprule
& \multicolumn{4}{c|}{Normal Light} & \multicolumn{4}{c}{Low Light} \\
\midrule
   Backbone &   NLL &   OPM &  DetPr &  LocA & NLL &   OPM &  DetPr &  LocA\\
\midrule
ResNet50 & 6.598 & 0.959 &  0.959 & 0.997 & 19.675 & 0.469 &  0.470 & 0.900 \\
    DETR & 7.149 & 0.985 &  0.987 & 0.990 &  6.834 & 0.917 &  0.919 & 0.978 \\
    Ours & 8.624 & 0.958 &  0.960 & 0.993 & 35.338 & 0.644 &  0.646 & 0.947 \\
\bottomrule
\end{tabular}
\label{tab:baseline}
\end{table*}

\textbf{Negative Log Likelihood (NLL):} Since any model that solves the heteroskedastic geospatial object tracking problem must output a probability distribution over the locations of tracked objects by definition, evaluating the quality of such a model's output using a likelihood-based evaluation metric is natural. In this work, all tracking models output Gaussian distributions over object location via the distribution over the Kalman tracker's latent state. We specifically evaluate the negative log likelihood of the Kalman tracker's marginal distribution over the object's location. Note that the individual heteroskedastic detection models also output probabilistic detections, and thus their performance can also be evaluated using negative log likelihood.

\textbf{Object Probability Mass (OPM):}
Within image plane tracking problems typically use object bounding boxes as the ground-truth objection location representation and within image place tracking models typically also output bounding boxes. This leads applying an Intersection-over-Union (IoU) metric to true and predicted bounding boxes as a natural performance metric. 

Under the assumption that it is sufficient for the probability distribution of an object's location to concentrate within the geospatial extent of the object, we can derive a related metric for the heteroskedastic geospatial tracking problem by computing the predicted probability mass that falls within the ground truth extent of an object. 

Given a predicted distribution with mean $\mu$ and covariance $\Sigma$ along with the ground truth location, orientation, and extent of the object being tracked, we 
compute the OPM using a Monte Carlo estimate of the underlying integral that defines the desired probability mass. We sample 1000 points from the predicted distribution and check what percentage of them fall inside the rectangle defining the object's physical extent. A visual depiction of the application of this metric is shown in Figure 1 in the supplemental material.

\textbf{Tracking Metrics:}
Like IoU, OPM has the useful propriety of being in the range $[0,1]$, with 1 being a perfect score. This allows us to use OPM as a drop-in replacement for IoU when using evaluation code for tracking metrics. In particular, we use evaluation code for the Higher Order Tracking Accuracy (HOTA) suite of tracking metrics \citep{hota}.

In the HOTA metrics, a bijective mapping between predicted and ground-truth objects is computed. The number of true positives ($|TP|$) is the number of matched pairs. The number of unmatched predictions is $|FP|$. The number of unmatched ground-truth objects is $|FN|$. A match is only considered valid if the similarity score (e.g. IoU) is greater than a given threshold $\alpha$. 
We can then compute the detection recall and precision ($DetRe$ and $DetPr$) for a single $\alpha$ value as shown below in Equation \ref{detpr} and \ref{detre}. These values are computed for a range of $\alpha$ values and the mean is reported.
\begin{align}
    \label{detpr} DetPr_{\alpha} &= \frac{|TP|}{|TP|+|FN|}\\
    \label{detre} DetRe_{\alpha} &= \frac{|TP|}{|TP|+|FP|}
\end{align}
In our data set, there is one ground-truth object and prediction per time step. Therefore, the only kind of error that can occur is that the detection results in a similarity score less than the threshold $\alpha$. This results in an unmatched ground truth object and an unmatched prediction, so $|FN|=|FP|$. This implies that $DetPr = DetRe$. We therefore chose to report $DetPr$ only. We reinterpret this metric to be the percent of timesteps with OMP greater than $\alpha$, averaged across a range of $\alpha$ values. That is, it measures how often the tracker is on track.

We additionally report the location accuracy (LocA) metric from the HOTA suite. LocA can be defined as shown below  in Equation \ref{loca} where $c$ is a valid ground-truth/prediction pair. Put simply, $LocA_{\alpha}$ is the average similarity score for pairs that are within the $\alpha$ threshold. That is, it measures the tracker's localization performance when it is on track. 
We again note that an individual heteroskedastic geospatial detection model can also be used to derive these metrics in the single object tracking case.
\begin{align}
    \label{loca} LocA_{\alpha} &= \frac{1}{|TP_{\alpha}|} \sum_{c \in TP_{\alpha }} OPM(c)
\end{align}

\begin{table*}[t]
\centering
\begin{minipage}{.5\linewidth}
\caption{Normal Light, Detectors vs. Tracker}
\label{lightsT_detectors}
\begin{tabular}{l|rr|rr}
\toprule
& \multicolumn{2}{c|}{NLL} & \multicolumn{2}{c}{OPM} \\
\midrule
   Backbone&  Detector &  Tracker  &  Detector  &  Tracker  \\
\midrule
ResNet50 &         6.361 &        6.598 &         0.803 &        0.959 \\
    DETR &         5.288 &        7.149 &         0.819 &        0.985 \\
    Ours &         7.873 &        8.624 &         0.701 &        0.958 \\
\bottomrule
\end{tabular}
\end{minipage}%
\begin{minipage}{.5\linewidth}
\caption{Low Light, Detectors vs. Tracker}
\label{lightsF_detectors}
\begin{tabular}{l|rr|rr}
\toprule
& \multicolumn{2}{c|}{NLL} & \multicolumn{2}{c}{OPM} \\
\midrule
   Backbone&  Detector &  Tracker  &  Detector  &  Tracker  \\
\midrule
ResNet50 &        11.329 &       19.675 &         0.197 &        0.469 \\
    DETR &         7.484 &        6.834 &         0.595 &        0.917 \\
    Ours &        17.489 &       35.338 &         0.308 &        0.644 \\
\bottomrule
\end{tabular}
\end{minipage}%
\end{table*}

\vspace{-1em}
\section{Results}\label{sec:results}
In this section we report the results of tracking experiments. We use the two data scenarios described in Section \ref{sec:dataset} (normal light vs. low light), the three backbone models described in Section \ref{sec:model} (ResNet50, DETR, and our custom model), and the evaluation metrics described in Section \ref{sec:experiments} (NLL, OPM, DetPr, LocA). The supplemental material includes a table of all results shown.

\textbf{Experiment 1 - Baseline Performance:}
We present baseline performance results for both data scenarios in Table \ref{tab:baseline}. These results evaluate the performance of the Kalman filter-based geospatial tracking (GST) model coupled with heteroskedastic geospatial detection (HGD) models based on each backbone. We label each model by the backbone used as the other components are identical across models. 

In terms of NLL, we see that results are fairly mixed apart from the result that the DETR backbone achieves much better NLL in the low light setting. In terms of OPM and the tracking metrics, we see that the DETR backbone outperforms the other backbones in both settings. It shows especially strong performance relative to the other models in the visually challenging low light setting. One possible explanation for this is that DETR is uniquely able to fuse global information by applying self-attention between all pixels in the final feature map. In the easier data scenario with normal light, we see that our more efficient model is competitive with the ResNet50 baseline in terms of OPM and tracking metrics. 

\textbf{Experiment 2 - Effect of Kalman Filtering:}
One benefit of using a late fusion approach is that each HGD model can function independently without the need of the centralized fusion process provided by the Kalman Filter. In Tables \ref{lightsT_detectors} and \ref{lightsF_detectors} we compare the average performance of the four HGD models compared to the performance of the baseline Kalman Filter model.

We can see that using the tracker leads to much better performance on the OPM metric. This suggests that the individual HGD models are producing higher uncertainty and larger covariance than the Kalman tracker output. After applying the tracker, the uncertainty is reduced and the covariance is significantly smaller. This can be seen in Figure \ref{fig:example_trajectories}. This results in a higher OPM score as the OPM metric is highly sensitive to the scaling of the covariance matrix. However, we see that  the average NLL across the four HGD models per backbone is actually better than the NLL provided by the Kalman tracker in all but one case. This suggests that from a predictive likelihood standpoint, the baseline Kalman tracker is actually under representing uncertainty. We address this issue in the next experiments.  

\begin{table*}
\centering
\begin{minipage}{.5\linewidth}
\caption{Calibration Results (NLL).}
\label{calib}
\begin{tabular}{l|rr|rr}
\toprule
& \multicolumn{2}{c|}{Normal Light} & \multicolumn{2}{c}{Low Light} \\
\midrule
   Backbone &  Uncalib.  &  Calib.  &  Uncalib.  &  Calib.  \\
\midrule
ResNet50 &                     6.598 &                   4.537 &                     19.675 &                    7.200 \\
    DETR &                     7.149 &                   5.465 &                      6.834 &                    4.803 \\
    Ours &                     8.624 &                   5.364 &                     35.338 &                    7.419 \\
\bottomrule
\end{tabular}
\end{minipage}%
\begin{minipage}{.5\linewidth}
\centering
\caption{Results with/without Kalman Filter tuning (NLL)}
\label{kf_train}
\begin{tabular}{l|rr|rr}
\toprule
& \multicolumn{2}{c|}{Normal Light} & \multicolumn{2}{c}{Low Light} \\
\midrule
   Backbone &  without  &  with  &  without  &  with  \\
\midrule
ResNet50 &                    4.537 &                   4.510 &                     7.200 &                    7.360 \\
    DETR &                    5.465 &                   5.980 &                     4.803 &                    5.002 \\
    Ours &                    5.364 &                   4.973 &                     7.419 &                    7.240 \\
\bottomrule
\end{tabular}
\end{minipage}%
\end{table*}

\begin{figure*}
\centering
\begin{subfigure}{.32\textwidth}
  \centering
  \includegraphics[width=\linewidth]{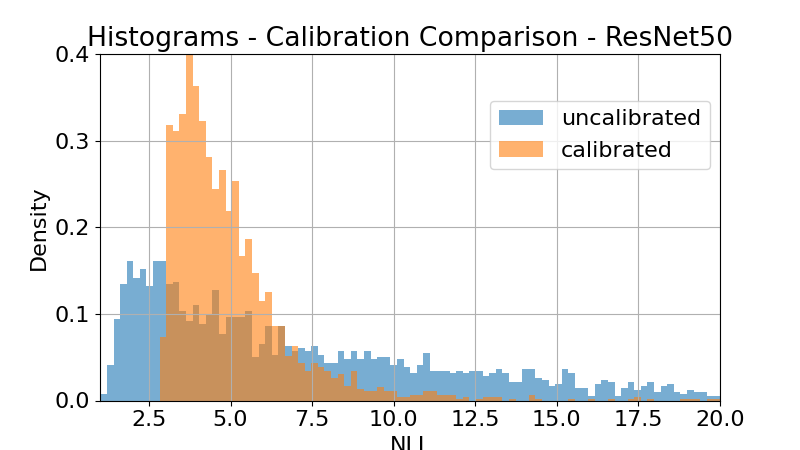}
\end{subfigure}%
\begin{subfigure}{.32\textwidth}
  \centering
  \includegraphics[width=\linewidth]{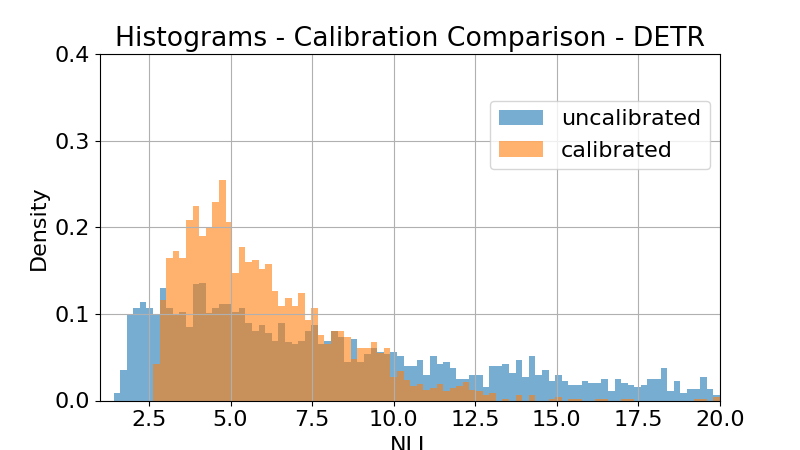}
\end{subfigure}
\begin{subfigure}{.32\textwidth}
  \centering
  \includegraphics[width=\linewidth]{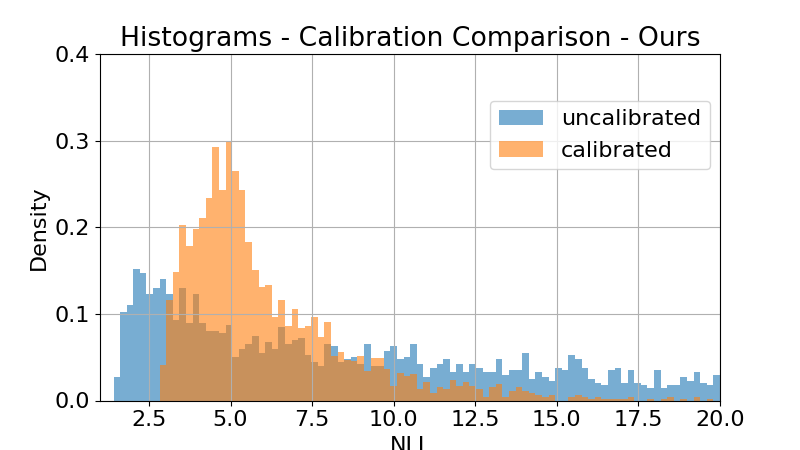}
\end{subfigure}
\caption{NLL Histograms pre- and post-calibration. Plots are generated using all test set data cases.}
\label{fig:calib_histograms}
\end{figure*}

\textbf{Experiment 3 - Post-Hoc Model Recalibration:}
As noted in the previous section, the end-to-end output of the Kalman tracker under represents uncertainty in a predictive log likelihood sense. As a first approach to addressing this issue, we consider post-hoc recalibration of the output of the individual HGD models. 
Specifically, we consider an affine transformation of the covariance matrix output by an HGD model:  $\Sigma' = a \Sigma + b I$. This transformation inflates the covariance matrix for parameters $a>1$ and $b>0$. 

To select the values of $a$ and $b$ we perform a grid search over $a \in [0.05 \dots 10]$ and $b \in [0 \dots 500]$. We assess the performance of each re-scaled distribution on a validation set and select the $a,b$ which minimize validation set NLL. 

The results of this calibration experiment are shown in Table \ref{calib}. We report the NLL of the Kalman Tracker using uncalibrated and calibrated input from the HGD models. Comparing the calibrated models to the uncalibrated models, we see consistent improvements in terms of NLL. We see particularly large improvements in the visually challenging low light scenario. On the easier normal light scenario, we see that our customized model slightly out performs the DETR baseline. However the ResNet50 baseline is still the strongest model in this scenario. 

We further generate histograms of the NLL values with and without calibration across all time steps in the test data set. These results are displayed for all backbones in Figure \ref{fig:calib_histograms}. We can see that the effect of calibration is to increase the minimum NLL somewhat for all models, while significantly reducing the upper tail of the NLL distribution. 

Overall we find that this form of calibration is an easy way to improve the NLL of the tracker. This is convenient as it incurs almost no additional runtime latency compared to the complexity of executing the HGD backbones.

\textbf{Experiment 4 - Kalman Tracker Fine Tuning:}
Inference in the Kalman tracker is fully differentiable. It is thus possible to learn while backpropagating the Kalman filter's NLL through both the parameters of the Kalman filter and the parameters of individual HGD models. We refer to this approach as Kalman filter fine tuning. 

In our implementation of the constant velocity Kalman tracker, the standard deviation of acceleration parameter is the only learnable parameter. However, this parameter is important as it determines how much the Kalman filter's uncertainty over its internal latent variables increases between observations, which has implications both for the uncertainty in the tracker's output and the degree of smoothness in the output track. 

Further, Kalman filter fine tuning allows the output heads and adapters inside the HGD models to adjust themselves to the tracking scenario (for computational complexity reasons, we do not consider full end-to-end training of the backbones). We present results for this setting in Table \ref{kf_train}. The results reported are using calibrated HGD models as described in the previous section. The reported NLLs are computed with respect to the Kalman tracker's output distribution. 

We see that applying Kalman filter fine tuning results in an improvement in out custom model on both data scenarios. Surprisingly, the DETR baseline sees worse NLL performance on both data settings. After calibrating and training through the Kalman Filter, our custom model is able to achieve better NLL score than DETR in the normal light scenario at six times lower latency. It is also able to close to within 0.5 nats of the performance achieved by the ResNet50 backbone in the normal light scenario at 2.5 times lower latency. Lastly, despite the fact that the DETR-based model does not improve in performance under Kalman filter fine tuning, it remains the best performing model in the low light scenario by a wide margin.

\vspace{-1em}
\section{Conclusions}\label{sec:conclusions}
In this paper, we focus on the geospatial object tracking problem using data from a distributed camera network. The goal is to predict an object's track in geospatial coordinates along with uncertainty over the object's location. We have presented a novel single-object geospatial tracking data set to support this work that includes high-accuracy ground truth object locations and video data from a network of four cameras captured under normal and low light conditions. 

We have presented a modeling framework for addressing this task that leverages a multi-observation Kalman filter-based tracker in conjunction with a set of independent heteroskedastic geospatial detection models. This framework is specifically motivated by a communication constrained version of the tracking problem where it is too expensive to centralize raw image data and thus late fusion of low dimensional representations is required. We present a custom detection backbone with significantly reduced prediction latency relative to state-of-the-art models and show that it can achieve strong performance in the normal lighting setting when tuned specifically to optimize tracking performance via a combination of post-hoc recalibration and Kalman filter based fine tuning. Our result show that a version of the proposed framework using the DETR backbone achieves superior performance in the low light setting, a finding that requires further exploration.

\bibliography{sources}
\end{document}


\onecolumn 
\maketitle

This Supplementary Material should be submitted as a separate file. Please do not append the Supplementary Material to the main paper. 

Fig. \ref{fig:pitt} and Eq \ref{eq:example} in the main paper can be cross referenced using \texttt{xr}. 

\appendix
\section{Additional simulation results}
Table~\ref{tab:supp-data} lists additional simulation results; see also \citet{einstein} for a comparison. 

\begin{table}[!h]
    \centering
    \caption{An Interesting Table.} \label{tab:supp-data}
    \begin{tabular}{rl}
      \toprule 
      \bfseries Dataset & \bfseries Result\\
      \midrule 
      Data1 & 0.12345\\
      Data2 & 0.67890\\
      Data3 & 0.54321\\
      Data4 & 0.09876\\
      \bottomrule 
    \end{tabular}
\end{table}

\section{Math font exposition}
\providecommand{\upGamma}{\Gamma}
\providecommand{\uppi}{\pi}
How math looks in equations is important:
\begin{equation*}
  F_{\alpha,\beta}^\eta(z) = \upGamma(\tfrac{3}{2}) \prod_{\ell=1}^\infty\eta \frac{z^\ell}{\ell} + \frac{1}{2\uppi}\int_{-\infty}^z\alpha \sum_{k=1}^\infty x^{\beta k}\mathrm{d}x.
\end{equation*}
However, one should not ignore how well math mixes with text:
The frobble function \(f\) transforms zabbies \(z\) into yannies \(y\).
It is a polynomial \(f(z)=\alpha z + \beta z^2\), where \(-n<\alpha<\beta/n\leq\gamma\), with \(\gamma\) a positive real number.

\bibliography{uai2023-template}